\documentclass[10pt,twocolumn,letterpaper]{article}

\usepackage[pagenumbers]{cvpr} 

\definecolor{cvprblue}{rgb}{0.21,0.49,0.74}
\usepackage[pagebackref,breaklinks,colorlinks,allcolors=cvprblue]{hyperref}
\usepackage{float}

\def\paperID{22} 
\def\confName{CVPR}
\def\confYear{2025}

\usepackage{xcolor}

\title{BG-HOP: A Bimanual Generative Hand-Object Prior}

\author{Sriram Krishna \qquad Sravan Chittupalli \qquad Sungjae Park \\
Carnegie Mellon University\\
\normalsize {\tt\{sskrishn, schittup, sungjae2\}@cs.cmu.edu}
}

\begin{document}
\maketitle


\begin{abstract}
In this work, we present \textbf{BG-HOP}, a generative prior that seeks to model bimanual hand-object interactions in 3D. We address the challenge of limited bimanual interaction data by extending existing single-hand generative priors, demonstrating preliminary results in capturing the joint distribution of hands and objects. Our experiments showcase the model's capability to generate bimanual interactions and synthesize grasps for given objects. We make code and models \href{https://github.com/sriramsk1999/bghop/}{publicly available}.
\end{abstract}    
\section{Introduction}
\label{sec:intro}

Hand-object interaction is fundamental for most tasks in daily life. Human manipulation frequently involves the simultaneous use of both hands; people routinely employ bimanual coordination for tasks like opening jars, typing on keyboards, cooking, and more. Consequently, modeling these interactions is crucial across numerous real-world domains, from AR/VR applications to robotic manipulation.

Although previous work such as Generative Hand-Object Prior (G-HOP) \cite{ye2024g} has advanced the modeling of single-hand-object interactions, the challenge of synthesizing and reconstructing bimanual interactions remains relatively unexplored. On the other hand, while there exist task-specific approaches for bimanual grasp synthesis \cite{shao2024bimanual} or reconstruction \cite{on20241st}, they lack a generalized framework for modeling the joint distribution of hands and objects in bimanual interactions.

This interaction isn't simply two independent hands operating separately; they function collaboratively, requiring models to capture the \textit{interdependencies} between both hands and the object. This understanding represents a critical step toward developing systems capable of interacting with the physical world with human-like dexterity.

Existing datasets predominantly focus on single-handed manipulation, creating a substantial gap in data pertaining to bimanual interaction. To address this limitation, we extend G-HOP to develop a bimanual Hand-Object Prior. Building on this foundation, we formulate a generative prior designed for bimanual interactions.

By addressing data scarcity challenges and developing a specialized generative model for bimanual manipulation, this work establishes a foundation for more sophisticated hand-object interaction systems.

In summary, our contributions are as follows:
\begin{itemize}
    \item We propose \textbf{BG-HOP}, the first generative prior designed for modeling bimanual hand-object interactions.
    \item We demonstrate that current dataset limitations make training such a prior from scratch infeasible, and present preliminary results by leveraging transfer learning from existing single hand-object interaction models.
    \item We present qualitative results of our approach on \textit{grasp synthesis} - generating physically plausible bimanual grasps for given object meshes.
\end{itemize}
\begin{figure*}[ht!]
    \centering
    \includegraphics[width=\textwidth]{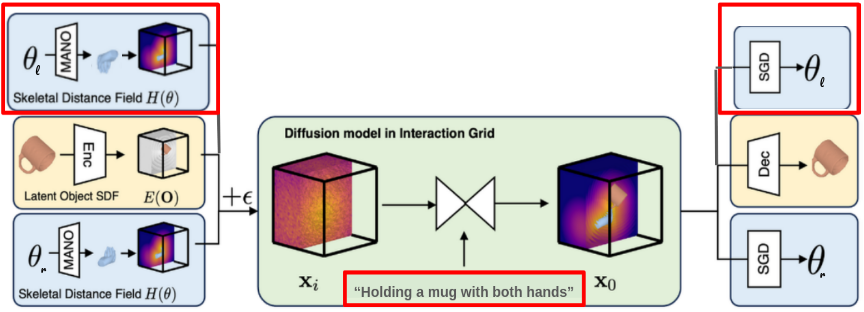}
    \caption{BG-HOP architecture. We extend G-HOP by concatenating the representations of both the left and the right hand along with the object latent code. We highlight the changes to the original architecture in \textcolor{red}{RED}.}
    \label{fig:bghop_architecture}
\end{figure*}

\section{Related Work}
\label{sec:related}

\textbf{Diffusion-Based Generative Models for Hand-Object Interaction:} 
Diffusion models \cite{ho2020denoising} have recently demonstrated significant advances in generative modeling in various domains, including image generation, video synthesis, and 3D asset creation. In the domain of 3D hand-object interactions, G-HOP \cite{ye2024g} establishes a framework for modeling the joint distribution of human hand parameters and object geometry. This unified distribution enables multiple downstream applications, including hand-object interaction reconstruction from videos and grasp synthesis. Our work extends this approach to capture the more complex dynamics of bimanual hand-object interactions.

\textbf{Hand-Object Interaction Synthesis:}
The synthesis of hand-object interactions is an active research area. Recent work has addressed bimanual interaction through diverse approaches: ArtiGrasp \cite{zhang2024artigrasp} synthesizes physically plausible bimanual grasping and object articulation using reinforcement learning based on hand pose references and object pose. Text2HOI \cite{cha2024text2hoi} and DiffH2O \cite{christen2024diffh2o} generate bimanual interactions from only object meshes and text descriptions. Shao et al. \cite{shao2024bimanual} focus specifically on synthesizing bimanual grasps for robotic manipulation. Although these approaches address various aspects of bimanual interaction, our work aims to develop a general-purpose generative prior. 

\textbf{Hand-Object Interaction Reconstruction:}
Recently, there has been significant progress in reconstructing hand-object interaction from images or videos. \cite{wu2024reconstructing} leverage transformers to predict human hand MANO\cite{romero2022embodied} parameters from a single RGB image. \cite{wu2024reconstructing, ye2024g, fan2024hold} further extracts the object mesh from the video, allowing reconstructing the interaction of the human hand and the object at the mesh level. While prior approaches provide promising results, in either synthesis or reconstruction, we aim to train a versatile model that can be used in both synthesis and reconstruction.

\textbf{Bimanual Hand-Object Interaction Datasets:}
Several datasets for bimanual hand-object interaction have emerged recently \cite{fan2023arctic, fu2024gigahands, zhan2024oakink2}. However, many 3D hand-object interaction datasets remain limited to single-hand scenarios \cite{chao2021dexycb, corona2020ganhand, liu2022hoi4d, yang2022oakink}. To address this constraint, we leverage transfer learning by fine-tuning from G-HOP \cite{ye2024g}, which demonstrates an understanding of single-hand object interaction. We explore various design choices to efficiently transfer these learned representations to the bimanual domain.
\section{Method}
\label{sec:method}

\subsection{Preliminaries}

Our work builds on the framework introduced by G-HOP \cite{ye2024g} for modeling hand-object interactions. This work formulates the problem as learning a joint distribution over (one) hand and an object. We extend this framework to the bimanual setting to model the distribution $p(O, H_l, H_r | C)$ where $C$ is the object category text prompt, $O$ represents the object, and $H_l, H_r$ represent the left and right hand, respectively.
 
\textbf{Representation:} Following G-HOP, objects are encoded as signed distance functions (SDF) compressed through a pre-trained vector quantized variational auto-encoder (VQ-VAE) \cite{van2017neural}. Hands are represented using \textit{interaction grids} — $n$-channel volumetric grids where each cell contains the distance to the corresponding hand joint. In the original formulation, these representations are concatenated and processed by a diffusion model that iteratively denoises them. The predicted object is recovered by decoding the denoised latent SDF through the VQ-VAE, while the hand parameters are recovered by optimizing the MANO \cite{romero2022embodied} parameters via gradient descent to minimize the distance to the predicted interaction grid. Please refer to \cite{ye2024g} for more details.

\begin{figure}[ht]
    \centering
    \includegraphics[width=\columnwidth]{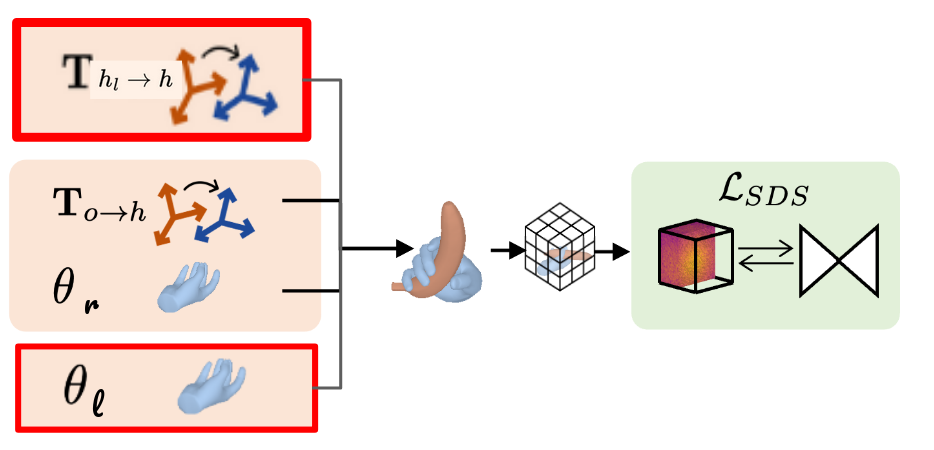}
    \caption{Object pose, left hand pose and articulations of both hands are arbitrarily initialized, and optimized to maximize the likelihood of interaction through Score Distillation Sampling (SDS). We highlight changes to the original optimization in \textcolor{red}{RED}.}
    \label{fig:biman_grasp_syn}
\end{figure}

\subsection{Bimanual Generative Hand-Object Prior}

To extend G-HOP to the bimanual setting, we modify the architecture to accommodate both left- and right-hand representations. The model processes a 3-tuple input consisting of the object representation and two hand interaction grids. As shown in Fig. \ref{fig:bghop_architecture}, our network architecture expands upon the single-hand pipeline by incorporating the representation for the left hand through concatenation. 

Processing occurs in the \textit{normalized right hand} coordinate frame - the object SDF is transformed into the coordinate frame of the right-hand, and the interaction grids of both hands are computed in the same frame. This unified reference frame enables the model to reason about the spatial relationships between the hands and the object.

However, this approach introduces a critical challenge: while the right-hand interaction grid represents its articulation, the left-hand interaction grid must encode both articulation and \textit{pose} relative to the right-hand. Thus, when sampling from the model, we must recover both the articulation and the relative spatial transformation of the left hand. We find that naively optimizing for both parameters simultaneously through gradient descent produces suboptimal results due to the ill-posed nature of this joint optimization.

To address this challenge, we introduce a Procrustes alignment procedure that decouples the optimization into sequential pose recovery and articulation recovery steps. First, we extract the 3D coordinates of the grid cells where the distance to each joint is minimized in the predicted distance field. Similarly, we extract corresponding coordinates from a neutral hand pose. With these two sets of $n=20$ corresponding 3D points (one for each joint), we solve the Procrustes problem to recover the rigid transformation $R|t$. Subsequently, we optimize for the left hand's articulation through gradient descent, following the same approach used for the right hand.

\begin{figure*}[t]
    \centering
    \setlength{\arrayrulewidth}{1.5pt}
    \begin{tabular}{ccc||ccc}
        \textbf{From-scratch} & \textbf{Semi-frozen} & \textbf{BG-HOP} & 
        \textbf{SDS Iter 0} & \textbf{SDS Iter 500} & \textbf{SDS Iter 1000} \\
        
        \includegraphics[width=0.13\textwidth, trim=10 10 10 10, clip]{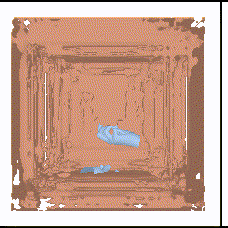} &
        \includegraphics[width=0.13\textwidth, trim=10 10 10 10, clip]{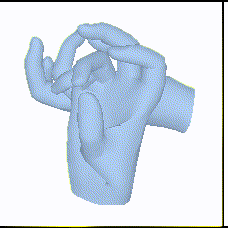} &
        \includegraphics[width=0.13\textwidth, trim=15 15 15 15, clip]{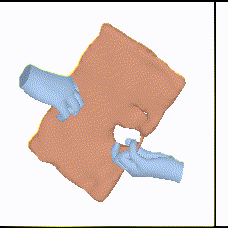} &
        \includegraphics[width=0.13\textwidth]{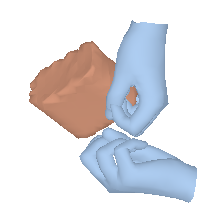} &
        \includegraphics[width=0.13\textwidth]{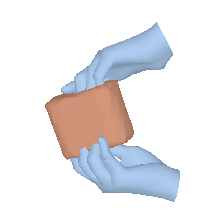} &
        \includegraphics[width=0.13\textwidth]{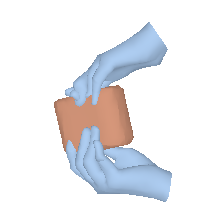} \\
        
        \multicolumn{3}{c||}{\textbf{Laptop}} & \multicolumn{3}{c}{\textbf{Potted Meat Can}} \\
        
        \includegraphics[width=0.13\textwidth, trim=10 10 10 10, clip]{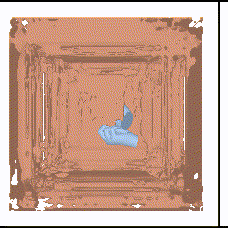} &
        \includegraphics[width=0.13\textwidth, trim=10 10 10 10, clip]{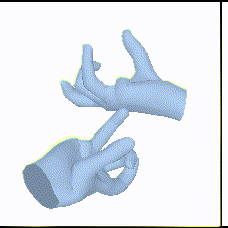} &
        \includegraphics[width=0.13\textwidth, trim=10 10 10 10, clip]{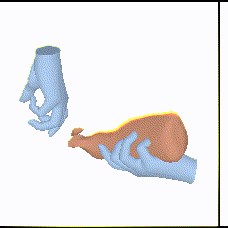} &
        \includegraphics[width=0.13\textwidth]{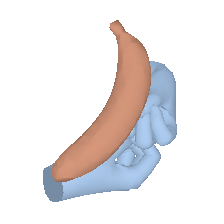} &
        \includegraphics[width=0.13\textwidth]{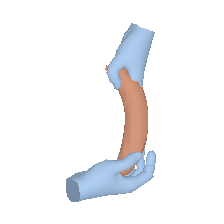} &
        \includegraphics[width=0.13\textwidth]{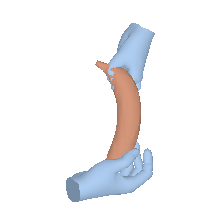} \\
        
        \multicolumn{3}{c||}{\textbf{Ketchup}} & \multicolumn{3}{c}{\textbf{Banana}} \\
        
        \includegraphics[width=0.13\textwidth, trim=10 10 10 10, clip]{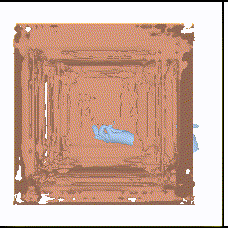} &
        \includegraphics[width=0.13\textwidth, trim=10 10 10 10, clip]{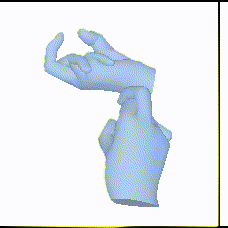} &
        \includegraphics[width=0.13\textwidth, trim=10 10 10 10, clip]{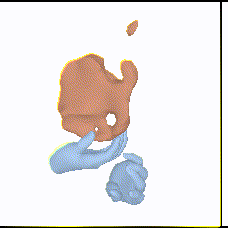} &
        \includegraphics[width=0.13\textwidth]{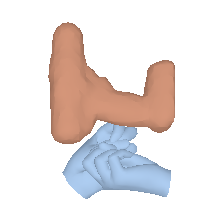} &
        \includegraphics[width=0.13\textwidth]{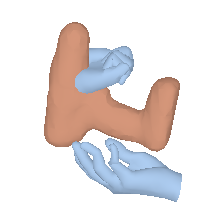} &
        \includegraphics[width=0.13\textwidth]{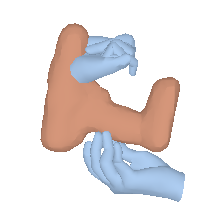} \\
        
        \multicolumn{3}{c||}{\textbf{Waffle Iron}} & \multicolumn{3}{c}{\textbf{Power Drill}} \\
    \end{tabular}
    \caption{Qualitative results of our BG-HOP framework. \textbf{Left}: Comparison of different training approaches for BG-HOP. While models trained from scratch and semi-frozen fail to converge, BG-HOP generates meaningful object geometry with appropriate hand positioning. \textbf{Right}: Progress of SDS-based optimization for grasp synthesis at different iterations. Starting with mean hand articulation and random poses, the model successfully positions objects in the right hand but sometimes struggles with left hand positioning and articulation.}
    \label{fig:combined_results}
\end{figure*}

\subsection{Bimanual Grasp Synthesis}

Given an object mesh as input, G-HOP synthesizes grasps through a test-time optimization approach with Score Distillation Sampling \cite{poole2022dreamfusion}. Specifically, the grasp is parameterized by the articulation of the hand $\theta$ and the relative transform between the hand and the object $T_{o\rightarrow h}$. These parameters are then optimized by maximizing the score function of the diffusion model.

We extend this approach to the bimanual setting as illustrated in Fig. \ref{fig:biman_grasp_syn}. Our method significantly expands the optimization parameter space to include both hands' articulations and their spatial relationships. Specifically, we optimize over four parameter sets: the articulations of both hands ($\theta_l, \theta_r$), the relative transformation between the object and the right hand ($T_{o\rightarrow h_r}$), and the relative transformation between the left hand and the right hand ($T_{h_l\rightarrow h_r}$). By optimizing all parameters simultaneously through SDS, we maximize the likelihood that the bimanual configuration reflects natural hand coordination patterns while maintaining appropriate contact points with the object geometry.

\subsection{Data}

Training a bimanual hand-object interaction model presents significant data challenges due to the limited availability of comprehensive bimanual datasets. We make use of the ARCTIC dataset \cite{fan2023arctic}, which provides high-fidelity motion capture data of bimanual interactions with various articulated objects. We implement a targeted preprocessing pipeline that extracts only frames where meaningful bimanual interaction occurs. Specifically, we select frames where a substantial percentage of vertices from both hand meshes are within a threshold distance $\epsilon$ of the object surface, indicating genuine contact.

As our model requires SDF representations, we develop a systematic approach to handle articulated objects. For each object, we compute SDFs at regular intervals across its complete articulation range, from fully closed to fully open states. We extract approximately 20,000 hand-object triplets across 11 distinct articulated objects.
\section{Experiments and Results}
\label{sec:results}

\subsection{Experimental Setup}

We examine BG-HOP through two primary tasks: bimanual interaction generation and bimanual grasp synthesis. To investigate the effectiveness of transfer learning from single-hand models, we train three variants:

\begin{itemize}
    \item \textbf{From scratch:} We train our model \textit{tabula rasa} without any initialization to establish a baseline.
    \item \textbf{Semi-frozen}: We initialize weights from G-HOP [Ye et al., 2024], modify the input/output layers to accommodate bimanual representations, and only train these new layers while keeping the pretrained weights frozen.
    \item \textbf{BG-HOP (full fine-tuning):} We initialize from G-HOP, adapt the architecture, and fine-tune the entire model end-to-end.
\end{itemize}

\subsection{Generation}

We visualize samples from the model in Fig. \ref{fig:combined_results} (Left). Our experiments reveal significant differences in convergence behavior across model variants. The \textbf{From-scratch} model fails to converge, demonstrating that the limited bimanual data available is insufficient for learning the complex joint distribution without prior knowledge. Similarly, the \textbf{Semi-frozen} approach fails to produce meaningful object representations, indicating that merely adapting the input/output layers is insufficient - the intermediate representations require substantial adaptation to the bimanual setting.

\textbf{BG-HOP} (with end-to-end fine-tuning) shows modest improvements over the other variants. In most samples, the model generates a reasonable approximation of object geometry, demonstrating effective transfer of object knowledge from G-HOP. 

However, significant challenges remain in coordinating bimanual poses. Despite our Procrustes alignment procedure, the left hand often exhibits positioning issues - either intersecting with the object or floating in free space without meaningful contact. Furthermore, once the hand is incorrectly positioned, the optimization process cannot converge and leads to infeasible articulations for the left hand. 

While end-to-end fine-tuning improves performance on objects present in the ARCTIC dataset, the model fails to generalize to text prompts describing objects outside the training distribution. This limitation highlights the need for more diverse bimanual interaction data spanning a wider range of object categories and interaction types.

\subsection{Grasp Synthesis}

In this section, we examine the ability of the model to synthesize grasps on a given object. Both hands are initialized with a mean articulation, while the left hand and the object are initialized with a random pose. Following this, we perform a test-time optimization with SDS loss and update the parameters controlling the position and configuration of the hands and the object.

We present results in Fig. \ref{fig:combined_results} (Right), on the HO3D dataset, whose objects are \textbf{unseen} during the bimanual finetuning. Similar to the results seen when sampling from the model in Fig. \ref{fig:combined_results} (Left), the model is usually able to meaningfully position the object such that it conforms to a meaningful grasp on the right hand. However, the performance is significantly degraded on the left hand, which may be positioned in unlikely locations and unnatural articulations (see \textbf{power drill} in Fig. \ref{fig:combined_results}) (Right).
\section{Conclusion}
\label{sec:conclusion}

Our experiments reveal several key limitations in the current approach:

\begin{itemize}
    \item \textbf{Data scarcity:} The limited availability of high-quality bimanual interaction data severely constrains model performance. While transferring from single-hand interaction models eases the amount of data required, a larger more diverse pool of data is still required to learn a truly general prior over bimanual hand-object interactions.
    \item \textbf{Hand coordination modeling:} The current framework struggles to model the complex spatial relationships between hands and objects. A more sophisticated architecture specifically designed to capture the interrelationships might yield better results.
    \item \textbf{Generalization:} The model's inability to generate plausible interactions for novel object categories indicates limited generalization. Future work could explore more robust text-to-3D conditioning mechanisms.
\end{itemize}

Despite these limitations, our work establishes a baseline for bimanual interaction modeling and identifies critical challenges that must be addressed to advance this field.
{
    \small
    \bibliographystyle{ieeenat_fullname}
    \bibliography{main}
}


\end{document}